\documentclass[letterpaper]{article}
\usepackage{aaai}
\usepackage{times}
\usepackage{helvet}
\usepackage{courier}

\usepackage{graphicx} 
\usepackage{algorithmic}
\usepackage{algorithm}

\newcommand{\sCOMMA} {\textsf{\footnotesize{s-COMMA}}}

\newcommand{\flatsCOMMA} {\textsf{\footnotesize{Flat s-COMMA}}}
\newcommand{\code}[1]{\texttt{\footnotesize{#1}}}
\newcommand{\Eclipse} {ECL$^{i}$PS$^{e}$}

\begin{document}
%
\title{Rewriting Constraint Models with Metamodels}

\author{Rapha\"el Chenouard$^{1}$ \and Laurent Granvilliers$^{1}$ \and Ricardo Soto$^{1,2}$\\
$^{1}$Universit\'e de Nantes, LINA, CNRS UMR 6241, France\\
$^{2}$Escuela de Ingenier{\'\i}a Inform\'atica\\ Pontificia Universidad
Cat\'olica de Valpara{\'\i}so, Chile\\
$\{$raphael.chenouard,laurent.granvilliers,ricardo.soto$\}$@univ-nantes.fr}

\maketitle

\begin{abstract}
An important challenge in constraint programming is to rewrite
constraint models into executable programs calculating the
solutions. This phase of constraint processing may require
translations between constraint programming languages, transformations
of constraint representations, model optimizations, and tuning of
solving strategies. In this paper, we introduce a pivot metamodel
describing the common features of constraint models including
different kinds of constraints, statements like conditionals and
loops, and other first-class elements like object classes and
predicates. This metamodel is general enough to cope with the
constructions of many languages, from object-oriented modeling
languages to logic languages, but it is independent from them. The
rewriting operations manipulate metamodel instances apart from
languages. As a consequence, the rewriting operations apply whatever
languages are selected and they are able to manage model semantic
information. A bridge is created between the metamodel space and
languages using parsing techniques. Tools from the software
engineering world can be useful to implement this framework.
\end{abstract}

\section{Introduction}
In constraint programming (CP), users describe properties of problems as
constraints involving variables. The computer system calls constraint
solvers to calculate the solutions. The automatic mapping from
constraint models to solvers is the key issue of this paper. The goal
is to develop middle software tools that are able to reformulate and
rewrite models according to solving requirements.

Modeling real-world problems requires high-level languages with many
constructions such as constraint definitions, programming statements,
and modularity features. In the recent past, a variety of languages
has been designed for a variety of users and problem categories. On
one hand, there are many modeling languages for combinatorial
problems such as OPL~\cite{VanHentenryckPPDP1999},
Essence~\cite{FrischIJCAI2007}, and MiniZinc~\cite{Nethercote2007} or
numerical constraint and optimization problems such as
Numerica~\cite{VanHentenryckBook1997} and
Realpaver~\cite{GranvilliersACM2006}. On the other hand, constraint
solving libraries have been plugged in computer programming
languages, for instance ILOG Solver~\cite{PugetSCIS1994},
Gecode~\cite{SchulteGecode2006}, and
\Eclipse{}~\cite{EclipseBook2007}. In the following, we will only
consider modeling languages as input constraint models. However,
computer programming languages can be chosen as targets of the mapping
process. Our aim is therefore to provide a many-to-many mapping tool that
is able to cope with a variety of languages.

Many constructions are shared among the different languages, in
particular the definitions of constraints. Other constructions are
specific such as classes in object-oriented languages or predicates in
logic languages. We propose to embed this collection of concepts in a
so-called metamodel, that is a model of constraint models. This pivot
metamodel describes the relations between concepts and it encodes in an
abstract manner the rules for constraint modeling. This is a
considerable improvement of our previous work~\cite{Chenouard2008}
which was restricted to a one-to-many mapping approach from a
particular modeling language. Moreover, the translations to obtain
\flatsCOMMA{} models were hand-coded and model structures are always
flattened like for FlatZinc models~\cite{Nethercote2007}. Previous model
transformations were also specific to \flatsCOMMA{} and its
structure (e.g. there is no object and no loop to manage). Our
pivot metamodel is independent of modeling languages and our approach
offers more flexibility in getting efficient executable models.

The rewriting process can be seen as a three-steps procedure. During
the first step, the user constraint model is parsed and a metamodel
instance is created. During the last step, the resulting program is
generated from a metamodel instance. These two steps constitute a
bridge between languages --- the grammar space --- and models --- the
model space. The middle step may implement rewriting operations over
metamodel instances, for instance to transform constraint
representations from an integer model to a boolean model. The main
interest is to manipulate concepts rather than syntactic
constructions. As a consequence, the rewriting operations can be
expressed with clarity and they apply whatever languages are chosen.

An interesting work is about the rule-based programming language
Cadmium~\cite{CadmiumICLP2006} combining constraint handling
rules~\cite{CHRBook2009} and term rewriting to transform constraint
models. The rewriting algorithm matches rules against terms in order
to derive some term normal forms. This approach provides a very clear
semantics to the mapping procedure and it addresses confluence and
termination issues. Considering metamodels allows one to reuse
metamodeling tools from software engineering. For instance,
ATL~\cite{KurtevATL2007} is a general rule-based transformation
language mixing model pattern matching and imperative programs, which
can be contrasted with term matching in
Cadmium. Kermeta~\cite{Kermeta} is a transformation framework allowing
to handle model elements using object-oriented programs. A benefit of
the model-driven approach is to directly manage typed model concepts
using the metamodel abstract description.

The remaining of this paper is organized as follows. Section 2
presents the general model-driven transformation framework underlying
this work. A motivating example using known CP languages is described
in Section 3. The pivot metamodel and rewriting operations are
presented in Section 4. Section 5 investigates some transformation
experiments on well-known CP models. Finally, Section 6 concludes the
paper and details some future work.

\section{Model Engineering Framework}\label{sec:overview}

\begin{figure}[htbp]
\begin{center}
\includegraphics[width=1\linewidth]
                {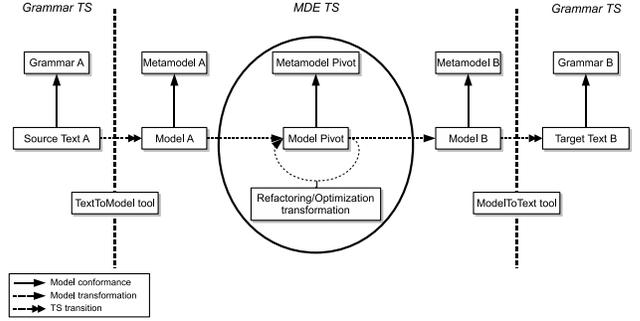}
\caption{Constraint model transformation process.}\label{fig:trans-process}
\end{center}
\end{figure}

A constraint model is a representation of a problem, written in a
language, and having a structure. Our purpose is to transform
solver-independent models to solver-dependent models. That may lead
\begin{itemize}
\item to change the representation of input models, namely the
  intrinsic constraint definitions, in order to improve the solving
  strategy,

\item to translate languages, from high-level modeling languages to
  low-level solver languages or computer programming languages, and

\item to modify model structures according to the capabilities of
  solvers, for instance to make a shift from object-oriented models to
  logic models based on predicates.
\end{itemize}
Managing representations supposes to specify constraint transformation
rules such as the equivalence of constraint formulations or constraint
relaxations. Translating languages requires to map concrete syntactic
elements. Manipulating structures deals with abstract modeling
concepts like objects or predicates. An important motivation is to
separate these different concerns. In particular, the equivalence of
constraint formulations is independent from the languages. This argues
in favour of a model technical space (MDE TS) gathering modeling
concepts and transformation rules and a grammar technical space
(Grammar TS) addressing the language issues, as shown in
Figure~\ref{fig:trans-process}.

In the grammar space, models are written in languages given
by grammars. In the model space, they are defined as relations
between elements that conform to metamodels. The elements are
instances of concepts described in metamodels, for example a
constraint $x+y=z$ deriving from some algebraic constraint concept.
The relations define links between concepts such as composition and
inheritance. That allows one to define complex elements, such as
constraint systems composed of collections of constraints.

The shift from languages to models can be implemented by parsing
techniques. Model \texttt{A} is created from the source user model
\texttt{A}. This model must conform to the user language metamodel, as
is required in the model space. As a consequence, metamodels of
languages ---modeling languages, constraint programming languages,
solver languages--- must be defined. The output \texttt{B} is generated
from model \texttt{B}. This model must conform to the metamodel of the
solver language. 

Model transformations are defined in the model space. The goal is to
transform model \texttt{A} reflecting the user constraint model to
model \texttt{B} associated to the solver.  As previously mentioned,
that requires to change model representations and structures. This
process can be done by rewriting operations manipulating concepts from
\texttt{A} to \texttt{B}. In order to share common concepts, we
propose to introduce the so-called pivot metamodel. The transformation
chain is then a three-steps procedure: a shift from model \texttt{A}
to the pivot model, the application of rewriting operations over the
pivot model, and a shift from the pivot model to model \texttt{B}.

In the following (Section 4), we will present the pivot metamodels and
model transformation operations. However, we will present first a
motivating example (next section) and discuss the requirements for
handling constraint models.

\section{A Motivating Example}\label{sec:CPMM}

Let us illustrate the transformation process on the social golfers
problem. The user model is written in the object-oriented modeling
language \sCOMMA{}. The output is a computer program written in
the constraint logic programming language \Eclipse{}.
This problem considers a group of $n=g\times s$ 
golfers that wish to play golf each week, arranged into $g$ groups of
$s$ golfers. The problem is to find a playing schedule for $w$ weeks
such that no two golfers play together more than
once. Figure~\ref{fig:scomma-golfers} and~\ref{fig:eclipse-golfers}
show the \sCOMMA{} model and the \Eclipse{} model for this problem,
respectively.

The \sCOMMA{} model is divided in a data file and a model file. The
data file is composed by an enumeration holding the golfer names, and
three constants to define the problem dimensions (size of groups,
number of weeks, and groups per week). The model file is divided into
three classes. One to model the groups, one to model the weeks and one
to arrange the schedule of the social golfers. The \code{Group} class
owns the \code{players} attribute corresponding to a set of golfers
playing together, each golfer being identified by a name given in the
enumeration from the data file. In this class, the code block called
\code{groupSize} (lines 14 to 16) is a constraint zone (constraint
zones are used to group statements such as loops, conditionals and
constraints under a given name). The \code{groupSize} constraint zone
restricts the size of the golfers group. The \code{Week} class has an
array of \code{Group} objects and the constraint zone
\code{playOncePerWeek} ensures that each golfer takes part of a unique
group per week. Finally, the \code{SocialGolfers} class has an array
of \code{Week} objects and the constraint zone \code{differentGroups}
states that each golfer never plays two times with the same golfer
throughout the considered weeks.

\begin{figure}[htbp!]
\begin{tiny}
\begin{verbatim}
//Data file
1. enum Name := {a,b,c,d,e,f,g,h,i};
2. int s := 3; //size of groups
3. int w := 4; //number of weeks
4. int g := 3; //groups per week

//Model file
1. main class SocialGolfers {
2.  Week weekSched[w];
3.  constraint differentGroups {
4.   forall(w1 in 1..w)
5.    forall(w2 in w1+1..w)
6.     forall(g1 in 1..g)
7.      forall(g2 in 1..g) {
8.        card(weekSched[w1].groupSched[g1].players intersect
          weekSched[w2].groupSched[g2].players) <= 1;
9.      }
10.   }
11. }
12. class Group {
13.  Name set players;
14.  constraint groupSize {
15.    card(players) = s;
16.  }
17. }
18.
19. class Week {
20.  Group groupSched[g];
21.   constraint playOncePerWeek {
22.    forall(g1 in 1..g)
23.     forall(g2 in g1+1..g) {
24.      card(groupSched[g1].players
          intersect groupSched[g2].players) = 0;
25.     }
26.   }
27. }
\end{verbatim}
\end{tiny}
\caption{An \sCOMMA{} model of the social golfers
  problem.}\label{fig:scomma-golfers}
\end{figure}

The generated \Eclipse{} model is depicted in
Figure~\ref{fig:eclipse-golfers}, which has been built as a single
predicate whose body is a sequence of atoms. The sequence is made of
the problem dimensions (lines 2 to 4), the list of integer sets
\code{L} (lines 6 to 7), and three nested loop blocks resulting from
the transformation of the three \sCOMMA{} classes (lines 9 to 36). It
turns out that parts of both models are similar. This is due to the
sharing of concepts in the underlying metamodels, for instance
constants, \code{forall} statements, or constraints. However, the
syntaxes are different and specific processing may be required. For
instance, the \code{for} statement of \Eclipse{} needs the
\code{param} keyword to declare parameters defined outside the current
scope, e.g. the number of groups \code{G}.

\begin{figure}[htbp!]
\begin{tiny}
\begin{verbatim}
1. socialGolfers(L):-
2.  S $= 3,
3.  W $= 4,
4.  G $= 3,
5.
6.  intsets(WEEKSCHED_GROUPSCHED_PLAYERS,12,1,9),
7.  L = WEEKSCHED_GROUPSCHED_PLAYERS,
8.
9.  (for(W1,1,W),param(L,W,G) do
10.  (for(W2,W1+1,W),param(L,G,W1) do
11.   (for(G1,1,G),param(L,G,W1,W2) do
12.    (for(G2,1,G),param(L,G,W1,W2,G1) do
13.      V1 is G*(W1-1)+G1,nth(V2,V1,L),
14.      V3 is G*(W2-1)+G2,nth(V4,V3,L),
15.      #(V2 /\ V4, V5),V5 $=< 1
16.    )
17.   )
18.  )
19. ),
20.
21. (for(I1,1,W),param(L,S,W,G) do
22.  (for(I2,1,G),param(L,S,W,G,I1) do
23.    V6 is G*(I1-1)+I2,nth(V7,V6,L),
24.    #(V7, V8), V8 $= S
25.  )
26. ),
27.
28. (for(I1,1,W),param(L,G) do
29.  (for(G1,1,G),param(L,G,I1) do
30.   (for(G2,G1+1,G),param(L,G,I1,G1) do
31.     V9 is G*(I1-1)+G1,nth(V10,V9,L),
32.     V11 is G*(I1-1)+G2,nth(V12,V11,L),
33.     #(V10 /\ V12, 0)
34.   )
35.  )
36. ),
37.
38. label_sets(L).
\end{verbatim}
\end{tiny}
\caption{The social golfers problem expressed in
  \Eclipse.}\label{fig:eclipse-golfers}
\end{figure}

The treatment of objects is more subtle since they must not
participate to \Eclipse{} models. Many mapping strategies may be
devised, for instance mapping objects to
predicates~\cite{SotoICTAI2007}. Another mapping strategy is used
here, which consists of removing the object-based problem
structure. Flattening the problem requires visiting the many classes
through their inheritance and composition relations.  A few problems
to be handled are described as follows. Important impacts on the
attributes may happen. For example, the \code{weekSched} array of
\code{Week} objects defined at line 2 of the model file in
Figure~\ref{fig:scomma-golfers} is refactored and transformed to the
\code{WEEKSCHED\_GROUPSCHED\_PLAYERS} flat list stated at line 6 in
Figure~\ref{fig:eclipse-golfers}. It may be possible to insert new
loops in order to traverse arrays of objects and to post the whole set
of constraints. For instance, the last block of for loops in the
\Eclipse{} model (lines 28 to 36) has been built from the
\code{playOncePerWeek} constraint  zone of the \sCOMMA{} model, but
there is an additional for loop (line 28) since the \code{Week}
instances are contained in the \code{weekSched} array. Another issue
is related to lists that cannot be accessed in the same way as arrays
in \sCOMMA{}. Thus, local variables (\code{V}$_{i}$) and the
well-known \code{nth} Prolog predicate are introduced in the
\Eclipse{} model.

\section{Pivot Model Handling}\label{sec:pivot-refactoring}


Our pivot metamodel
has been defined to catch most modeling needs that occur in constraint
modeling languages. Then, pivot models are managed with several
refining transformations, where each transformation identifies a clear
refining process, namely structure modifications (e.g. removal of
object variables) or model optimization.


\subsection{Pivot Metamodel}

\begin{figure*}[htbp]
\begin{center}
  \includegraphics[width=0.7\linewidth]
 {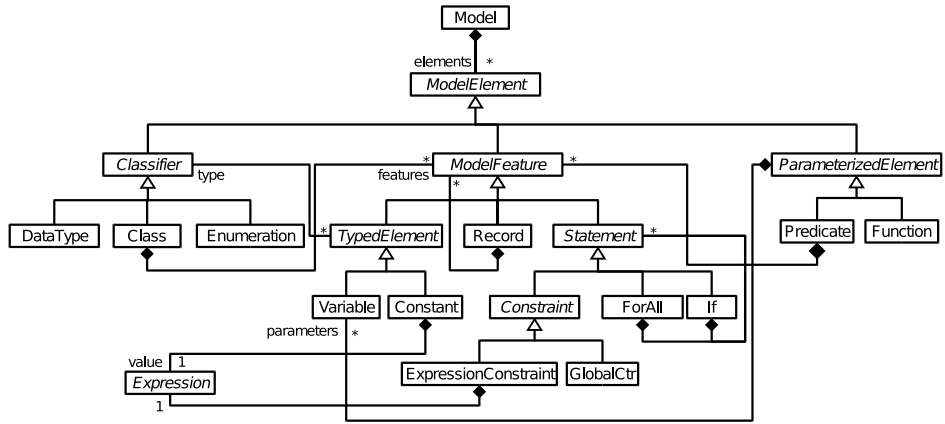}
\caption{Representation of variables and problem structures in the
  pivot metamodel.\label{fig:pivot-mm}}
\end{center}
\end{figure*}

Figure~\ref{fig:pivot-mm} depicts an extract of our pivot structure
metamodel in a simplified UML Class diagram formalism. Italic font is
used to denote abstract concepts. The root concept is 
\code{Model} which contains all entities. Three abstract concepts
inherit from the abstract class \code{ModelElement}:
\begin{itemize}
\item \code{Classifier} represents all types than can be used to
  define variables or constants: 
  \begin{itemize}
  \item \code{DataType} corresponds to common primitive data types used in
    CP, namely Boolean, Integer and Real.
  \item \code{Enumeration} is used to define symbolic types, i.e. a set
    of symbolic values defined as \code{EnumLiteral} (not defined here
    to keep the figure readable), e.g. \code{enum
      Name:=\{a,b,$\ldots$\}}, line 1 of data file in
    Figure~\ref{fig:scomma-golfers}.
  \item \code{Class} is similar to the object-oriented concept of class,
    but defined in a CP context \cite{SotoICTAI2007}, i.e. a class
    definition is composed of variable or constant definitions and also
    constraints and other statements. Thus, a \code{Class} has a set of
    features being instances of \code{ModelFeature}.
  \end{itemize}
\item \code{ModelFeature} corresponds to the instance concepts defined
  within a model. It is also divided in three concepts:
  \begin{itemize}
  \item \code{Record} relates to non-typed instances being composed of
    a collection of elements, such as tuples. To cover a broader range
    of record definitions, we define a composition of
    \code{ModelFeature} instances.
  \item \code{TypedElement} is an abstract concept corresponding to
    typed constraint model elements. Thus, it has a reference to a
    classifier. The concept of array variable is not distinguished
    from variable, but array can be represented using a sequence of
    sizes, corresponding to each dimension of an array (more than two
    dimensions are allowed). Theses sizes are expressed as
    \code{Expression} instances.
    \begin{itemize}
    \item \code{Variable} has an optional \code{Domain} definition
      (not shown here) restricting values belonging to the associated
      type. Three concepts of \code{Domain} are taken into account:
      intervals, sets and domains defined as an expression.
    \item \code{Constant} concept is for constants having a type and a
      fixed value.
    \end{itemize}
  \item \code{Statement} is used to represent all the other features
    that may occur in a \code{Model} or a \code{Class}:
    \begin{itemize}
    \item \code{Constraint} is the abstract constraint concept having
      two sub-concepts. \code{ExpressionConstraint} stands for
      constraints built inductively from terms and
      relations. \code{GlobalCtr} handles global constraints defined by
      a name and a list of parameters.
    \item \code{ForAll} defines a loop mechanism over constraints and
      other statements. It has an iterating variable which is local to
      the loop.
    \item \code{If} obviously defines a conditional statement. It is
      composed of an \code{Expression} corresponding to the boolean
      test and two sets of statements corresponding to the statements
      to take into account according to the test evaluation. The
      second set of statements is optional if no alternative to the
      true evaluation of the test is defined.
    \end{itemize}
  \end{itemize}
\item \code{ParameterizedElement} defines concepts having a list of
  parameters and not being a classifier neither an instance of a
  \code{ModelFeature}:
  \begin{itemize}
  \item \code{Predicate} represents logical predicates in a
    model as in \Eclipse{}. Predicates have parameters and a
    body composed of a sequence of \code{ModelFeature}, such as
    variable definitions or constraint statements.
  \item \code{Function} represents user-defined functions stated in a
    model. It contains also a body, but it is based on a statement
    used to compute a result.
  \end{itemize}
\end{itemize}

\begin{figure*}[htbp]
\begin{center}
  \includegraphics[width=1\linewidth]
 {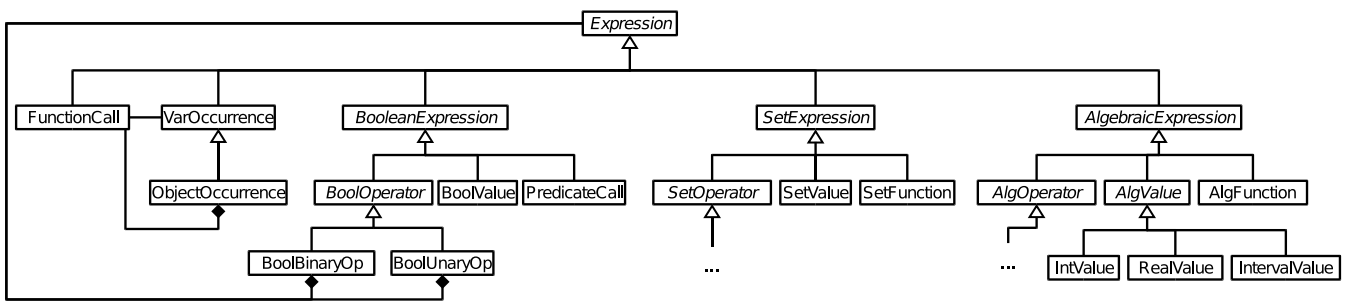}
\caption{Representation of expressions used to define constraint
  expression in the pivot metamodel.\label{fig:pivot-expr}}
\end{center}
\end{figure*}

The notion of expression is ubiquitous in CP. The related concepts of
our metamodel are detailed in Figure~\ref{fig:pivot-expr}. They
represent all the entities occurring in first-order formulas made from
variables, terms, relations, and connectives.
The concept \code{Expression} is abstract and is used as super class
for all kinds of expressions:
\begin{itemize}
\item \code{FunctionCall} is used to refer to an already defined
  \code{Function} and contains a list of parameters defined as
  \code{Expression}.
\item \code{VarOccurrence} is used to refer to already defined
  instances: records, variables or constants. It is only composed of a
  reference to the corresponding instance declaration and to a list of
  optional indexes to handle arrays. It is specialized in
  \code{ObjectOccurrence} in order to express the navigation path to an
  object attribute (e.g. \code{groupSched[g1].players}, line 8 in
  Figure~\ref{fig:scomma-golfers}). Variable occurrences are not
  classified according to their declaration type in one of the three
  expression types inheriting from \code{Expression} in order to avoid
  multiple declaration of the same concept, while requiring type
  inference mechanism.
\item \code{BooleanExpression} is used to specify boolean concepts
  occurring in expressions:
  \begin{itemize}
  \item \code{BoolValue} represents the terms \code{true} and
    \code{False}.
  \item \code{PredicateCall} corresponds to the call of a predicate
    with its list of parameters. Thus, it is composed of a reference
    to a \code{Predicate} with a list of parameters defined as
    \code{Expression} in order to allow at the same time
    \code{VarOccurrence} and evaluable expressions, such as $1$, $x$ or
    $x+1$.
  \item \code{BoolOperator} is an abstract concept having a name
    representing the symbol of well-known operators. It is
    specialized in the two common types of operators:
    \begin{itemize}
    \item \code{BoolUnaryOp} corresponds to the negation operator and
      has an operand corresponding to an \code{Expression}, since it
      can be a boolean expression, but also a variable. In the
      following, operands of all operators will be defined as a
      composition of \code{Expression}.
    \item \code{BoolBinaryOp} corresponds to the several common binary
      operators returning a boolean value, such as: $\Leftrightarrow$,
      $\rightarrow$, and, or, $=$, $\neq$, $\leq$, $\geq$, $<$, $>$.
    \end{itemize}
  \end{itemize}
\item \code{SetExpression} defines the main constructs available to
  deal with sets within expressions:
  \begin{itemize}
  \item \code{SetValue} corresponds to a set of value occurrences, such
    as $\{1,2,3\}$. To tackle various contents as set elements
    (e.g. $\{1,x+1\}$), it is composed of a list of elements conformed
    to \code{Expression}.
  \item \code{SetFunction} corresponds to the call of known
    functions over sets, such as the cardinality function.
  \item \code{SetOperator} is specialized only in \code{SetBinaryOp}
    since no unary operator is commonly used on sets. For instance,
    intersection, union and difference are available.
  \end{itemize}
\item \code{AlgebraicExpression} defines the numerical expressions:
  \begin{itemize}
  \item \code{AlgValue} is abstract and represents the three main
    concepts of values in numerical expressions: \code{IntValue} for
    integer values, \code{RealValue} for real number values and
    \code{IntervalValue} for interval values such as $[-1,1]$.
  \item \code{AlgFunction} corresponds to the call of a well-known
    function over numbers, such as trigonometric functions.
  \item \code{AlgOperator} refers to the common operators used in algebraic
    expressions: \code{AlgUnaryOp}($-$, $+$) and \code{AlgBinaryOp}
    ($+$, $-$, $*$, $/$ and \^{}).
  \end{itemize}
\end{itemize}

Our pivot metamodel has been defined to fit with most modeling needs
in CP, but also to fit with the metamodel of CP languages.
Thus, some simplifications have been done to ease transformations
such as the \code{VarOccurrence} concept which directly inherits from
\code{Expression}. Indeed, variable occurrences can be typed in
expressions (i.e. boolean, set or algebraic), but we define only one
to avoid redundancies.



\subsection{Pivot model refactoring}




Model transformations are implemented as rewriting operations over
pivot models.

 For sake of clarity, we will present a few operations using an imperative
pseudo-code style, while specific transformation languages are used in
practice. The main interest given by the concept hierarchy is to provide
navigation mechanisms through models. For instance, it is immediate to
iterate over the set of variables of a constraint, since this information
is gathered in the corresponding abstract constraint concept (see e.g.
Algorithm 2). It is therefore possible to manipulate models globally, which is
very powerful.

\subsubsection{Object flattening}

This refactoring step replaces object instances, namely variables
whose type is a class, by all elements defined in the class definition
(variable, constants, constraints and other statements).
In order to prevent name conflicts, named elements are prefixed 
with the name of object instances.

\begin{algorithm}
\caption{ Transforming and removing object variables and class definitions}
{\bf objectRemoval}(m : \code{Model})\\
: \code{Model}
\label{rule:object}
\algsetup{indent=2em}
\begin{algorithmic}[1]
   \STATE \textbf{let} res : \code{Model} 
   \FORALL{o \textbf{in} m.elements}
     \IF{$is\_var$(o) \textbf{and} $is\_class$(o.type)}
       \STATE res.insert(\textbf{flatten}(o,o.type.features))
     \ELSIF{{\bf not} $is\_class(o)$}
       \STATE res.elements.insert(o)
     \ENDIF
   \ENDFOR
   \RETURN res
\end{algorithmic}
\vspace{0.2cm}
{\bf flatten}(o : \code{Variable}, features : {\bf Set} of
\code{ModelFeature})\\
: {\bf Set} of \code{ModelFeature}
\algsetup{indent=2em}
\begin{algorithmic}[1]
  \STATE \textbf{let} res : {\bf Set} of \code{ModelFeature} =
  $\emptyset$ 
  \FORALL{f \textbf{in} features}
    \IF{$is\_var$(f) and \textbf{not} $is\_class$(f.type)}
      \STATE \textbf{let} v : \code{Variable}
      \STATE v $\leftarrow$ duplicate(f)
      \STATE v.name = o.name + '\_' + v.name
      \STATE res.insert(v)
    \ELSE
    \STATE ...
    \ENDIF
  \ENDFOR
  \RETURN res
\end{algorithmic}

\end{algorithm}

This refactoring transformation can be expressed in terms of a brief
pseudo-code algorithm as shown in Algorithm~\ref{rule:object}. The
\code{ObjectRemoval} function processes a source model by
iterating on all its elements (line 2). If object instances are
detected (line 3), then the function \code{flatten} is called and its
result is added to the output model elements (line 4). Instances not
being a \code{Class} definition are duplicated in the output model
(line 5,6), while \code{Class} definitions are removed. In the
\code{flatten} function every feature given as parameter is cloned and
added to the resulting set of \code{ModelFeature}. In the case
of a variable (and also constants), its name is concatenated to the
object variable name (line 6). Figure~\ref{fig:obj-rem} depicts the
result of the transformation on the social golfers example previously
presented.

\begin{figure}[!htpb]
\begin{small}
\code{class SocialGolfers \{ Week weekSched[w];$\ldots$\}}\\
\code{class Week          \{ Group groupSched[g];$\ldots$\}}\\
\code{class Group         \{ Name set players;$\ldots$\}}\\
$\quad\Rightarrow$  \code{Name set weekSched\_groupSched\_players[g*w];}
\end{small}
\vspace{-0.4cm}
\caption{Applying the object flattening transformation on the social
  golfers example using \sCOMMA{} syntax.\label{fig:obj-rem}}
\end{figure}

Arrays of objects and expressions refactoring are not presented here
to keep the algorithm simple. As mentioned at the end of Section 3,
in the case of object arrays, we must transfer their size to their
attributes and a loop statement has to be introduced to iterate on their
\code{Statement} instances. Within expressions, instances of
\code{VarOccurrence} may just be updated with the declaration of the
new flat variables.


\subsubsection{Alldifferent removal}
Since global constraints are not handled by every solver, there is a
motivation to reformulate them or to generate relaxations.
We consider here, the well-known global constraint
\code{alldifferent}($x_1,...,x_n$). We assume that the domain of each
$x_i$ varies from $1$ to $n$ to ease the definition of the two last
algorithms. We propose three possible transformations:
\begin{itemize}
\item Generating a set of disequalities as shown in
  Algorithm~\ref{rule:alldiff1}. For all variable combinations (line
  2,3), a constraint is generated and added to the result (line 6).
\begin{algorithm}
\caption{Transforming alldifferent to a set of disequalities}
\label{rule:alldiff1}
{\bf AllDiffToDisequalities}(c : \code{GlobalConstraint})\\
: {\bf Set} of \code{Constraint}
\begin{algorithmic}[1]
  \STATE \textbf{let} res : {\bf Set} of \code{Constraint} =
  $\emptyset$ 
  \FORALL{$i$ \textbf{in} $1 ..$c.parameters.size()}
    \FORALL{$j$ \textbf{in} $i+1 ..$c.parameters.size()}
      \STATE let x : \code{Variable} =  c.parameter[i]
      \STATE let y : \code{Variable} =  c.parameter[j]
     \STATE res.insert({\bf new}
      \code{Constraint}(x $\neq$ y))
    \ENDFOR
  \ENDFOR
  \RETURN res
\end{algorithmic}
\end{algorithm}
\item Generating a relaxation as shown on
  Algorithm~\ref{rule:alldiff2}. Only one constraint is created
  (line~3) assessing that the sum of all variable values is equal to $n(n+1)/2$.
\begin{algorithm}
\caption{Generating alldifferent relaxations}
\label{rule:alldiff2}
{\bf AllDiffToRelaxation}(c : \code{GlobalConstraint})\\
: \code{Constraint}
\begin{algorithmic}[1]
  \STATE {\bf let} n : Integer = c.parameters.size()
  \STATE {\bf let} sum : \code{Expression} = $\sum_{i=1}^{n}$
  c.parameters[$i$]
  \RETURN {\bf new} \code{Constraint}(sum = n(n+1)/2)
\end{algorithmic}
\end{algorithm}
\item Generating a boolean version as shown on
  Algorithm~\ref{rule:alldiff3}. In this case, we define a new matrix of
  boolean variables (line 2,3,4), where $b[i,j]$ being true means $x_i$ has
  value $j$. Line 7 checks that only one value per variable is
  defined. Line 10 ensures that two variables have different values.
\begin{algorithm}
\caption{Reformulating alldifferent into a boolean model}
\label{rule:alldiff3}
{\bf AllDiffToBoolean}(c : \code{GlobalConstraint})\\
: {\bf Set} of \code{ModelFeature}
\begin{algorithmic}[1]
  \STATE \textbf{let} res : {\bf Set} of \code{Constraint} = $\emptyset$ 
  \STATE {\bf let} n : Integer = c.parameters.size()
  \STATE {\bf let} m : Integer = card(c.parameters.domain)
  \STATE {\bf let} b[n,m] : Boolean
  \STATE res.insert(b)
  \FOR{$i$ {\bf in} $1..n$}
    \STATE res.insert({\bf new} \code{Constraint}($\sum_{j=1}^{m}$
    b[$i$,$j$] = 1))
  \ENDFOR
  \FOR{$j$ {\bf in} $1..m$}
    \STATE res.insert({\bf new} \code{Constraint}($\sum_{i=1}^{n}$
    b[$i$,$j$] = 1))
  \ENDFOR
\end{algorithmic}
\end{algorithm}
\end{itemize}



\section{Experiments}\label{sec:experiments}

The presented architecture has been implemented with three 
tools and languages: KM3~\cite{Jouault2006KM3} is a metamodel language,
ATL~\cite{KurtevATL2007} is a declarative rule language to describe model
transformations and TCS~\cite{Jouault2006TCS} is a declarative
language based on templates to define the text to model and model to
text transitions. These MDE tools allow us to choose the refactoring
steps to apply on pivot models in order to keep supported
structures of the target metamodel.

We have carried out a set of tests in order to analyze the performance
of our approach. We used five CP problems: Social Golfers, Engine Design,
Send+More=Money, Stable Marriage and N-Queens. The first experiment
evaluates the performance in terms of translation time, and the second
one was done to show that the automatic generation of solver files does
not lead to a loss of performance in terms of solving time. The benchmarking
study was performed on a 2.66Ghz computer with 2GB RAM running Ubuntu.

\begin{table}[htpb!]
\begin{scriptsize}
\begin{center}
\begin{tabular}{|l|c|c|c|c|c|c|c|}
\hline
Problems &   sC & s-to-P & Object & Enum & P-to-E & Total & Ecl\\
        &  Lines        & (s)    & (s)    & (s)  & (s)    & (s)   &  Lines\\
\hline
\hline
Golfers     & 31  & 0.276 & 0.340 & 0.080 & 0.075 & 0.771 & 37\\
\hline
Engine      & 112 & 0.292 & 0.641 & 0.146 & 0.087 & 1.166 & 78\\
\hline
Send        & 16  & 0.289 & 0.273 & -     & 0.089 & 0.651 & 21\\
\hline
Marriage    & 46  & 0.330 & 0.469 & 0.085 & 0.067 & 0.951 & 26\\
\hline
10-Q        & 14  & 0.279 & 0.252 & -     & 0.033 & 0.564 & 12\\
\hline
\end{tabular}
\end{center}
\end{scriptsize}
\caption{Times for complete transformation chains of several classical problems.}\label{tab:bench1}
\end{table}

In the first experiment we test the \sCOMMA{} (sC) to \Eclipse{} (Ecl)
translation. Table~\ref{tab:bench1} depicts the results. The first column
gives the problem names. The second column shows the number of lines of the
\sCOMMA{} source files. The following columns correspond to the time of atomic
steps involved in the transformation (in seconds): transformations
from \sCOMMA{} to Pivot (s-to-P) (corresponds to Source Text A
to Model Pivot in Figure~\ref{fig:trans-process}), object flattening (Object),
enumeration removal (Enum), and transformations from Pivot to \Eclipse{} (P-to-E)
(corresponds to Model Pivot to Target Text B in Figure~\ref{fig:trans-process}).
The next column details the total time of the complete transformation, and the
last column depicts the number of lines of the generated \Eclipse{} files.

The results show that the text processing phases (s-to-P and
P-to-E) are fast, but we may remark that the given problems are
concisely stated (maximum of 112 lines). The transformation \sCOMMA{}
to pivot is slower than the transformation pivot to \Eclipse{}.
This is explained by the refactoring phases performed on the pivot that
reduce the number of elements to handle the pivot to \Eclipse{} step.
The composition flattening is the more expensive phase. In particular,
the Engine problem exhibits the slowest running time, since it contains
several object compositions. In summary, considering the whole set of
phases involved, we believe the results show reasonable translation
times.

In the second experiment we compare the \Eclipse{} files automatically
generated by the framework with native \Eclipse{} files written by
hand (see Table~\ref{table:modelsize}). We consider the solving time
and the lines of each problem file. The data of the native models is
first given. We then introduce generated files where the loops have
not been unrolled (avoiding this phase the size of generated solver
files is closer to the native ones). In this case, the solving times
of both types of files are almost equivalent. At the end, we consider
problems including the loop unrolling phase (Flat). This process leads
to a considerable increase of model sizes. Only the solving time of the
flat 20-Queens and 28-Queens problems are impacted (about 0.4 and 7
seconds). 
This may be explained by the incremental propagation algorithm
commonly implemented in CLP languages. We may suppose that a propagation
happens each time a constraint is added to the constraint store. If a
for statement is not interleaved with propagation, i.e. it is considered
as one block, then only one propagation step is required. This is not
the case if loops are unrolled, leading to one propagation for each
individual constraint. It results in a slow-down.
This negative impact in terms of solving time demonstrates the need for keeping
the structure of target models (e.g. not unrolling loops) instead of building a
flat model.

\begin{table}
\begin{scriptsize}
\begin{center}
\begin{tabular}{|l||c|c|c|c|c|c|}
\hline
&\multicolumn{2}{c|}{Native}&\multicolumn{2}{c|}{Generated}&\multicolumn{2}{c|}{Generated (Flat)}\\
\cline{2-7}
Problems&solve(s)&Lines&solve(s)&Lines&solve(s)&Lines\\
\hline
\hline
Golfers            &0.21 &28   &0.21 &31  &0.22 &276 \\
\hline
Marriage           &0.01 &42   &0.01 &46  &0.01 &226 \\
\hline
20-Q               &4.63 &11   &4.65 &12  &5.02 &1162 \\
\hline
28-Q               &80.73 &11  &80.78 &12 &87.73 &2284 \\
\hline
\end{tabular}
\end{center}
\end{scriptsize}
\caption{Solving times and model sizes of native and generated files\label{table:modelsize}}
\end{table}

\section{Related Work}\label{sec:related}

Model transformation is a recent research topic in CP.
Just a few CP model transformation approaches
have been proposed. The solver-independent architecture is
likely to be the nearest framework to our approach, for
instance, MiniZinc (and Zinc), Essence and \sCOMMA{}.

MiniZinc is a high-level constraint modeling language
allowing transformations to \Eclipse{} and Gecode
models. These mappings are implemented by means of Cadmium.
The translation process involves an
intermediate model where several MiniZinc constructs are replaced by
simplified or solver-supported constructs. This facilitates the
translation to get a solver model.

Essence is another language involving model
transformations. Its solver-independent platform that allows to map
Essence models into \Eclipse{} and Minion~\cite{GentECAI96}. A model
transformation system called Conjure~\cite{FrischIJCAI2005} is included
in the framework, which takes as input an Essence specification and refines
it to an intermediate language called Essence'. The translation from
Essence' to solver code is currently performed by java translators using the
tool Tailor. 

\sCOMMA{} is an object-oriented language, supported by
a solver-independent platform where solvers can be mapped to
\Eclipse{}, Gecode/J, RealPaver, and GNU
Prolog~\cite{DiazSAC2000}. The language also involves an intermediate
model called \flatsCOMMA{} to facilitate the translation. Hand-written
translators and MDE-translators have been developed to translate
a \flatsCOMMA{} model in the target solver model.

Our approach can be seen as a natural evolution of this
solver-independent architecture. Two major advantages arise. (1) In
the aforementioned approaches just one modeling language can be used
as the source of the transformation, in our framework many modeling
language can be plugged as the source. We believe this enables
flexibility and provides freedom to the modelers. (2) In the
\sCOMMA{}, MiniZinc, and Essence transformation  processes, the
refactoring steps (e.g. enumeration removal, loop and set
unrolling) are always applied. This makes the
structure of the solver file completely different from the original
model. In our framework we focus on generating optimized models while
trying to maintain as much as possible the original structure of the
source model.
We believe that
keeping the source modeling structures into target models, then
improve their readability and understanding.

\section{Conclusion and Future Work}

In this paper, we have presented a new framework for constraint model
transformations. This framework is supported by an MDE approach
and a pivot metamodel that provides independence and flexibility
to cope with different languages. The transformation
chain involves three main steps: from the source to the pivot
model, refining of the pivot model and from the pivot model to
the target. Among others, an important feature of this chain
is the modularity of mode transformations and that the hard
transformation work (refactoring/optimization) is always performed
over the pivot. This makes the transformations 
from/to pivot simpler, and as a consequence the integration of new
languages to the architecture requires less effort.

In a near future, we intend to increase the number of CP languages our
approach supports. We also want to define more pivot refactoring
transformations to optimize and reformulate models. Another major
outline for future work is to improve the management of complex CP
models transformation chains, which is not investigated in this
paper. The order in which refactoring steps are applied and which
refactoring step to apply can be automated. However, we may
investigate how to qualify models and transformations according to
the pivot and target metamodels.

\bibliographystyle{aaai}
\bibliography{sara}

\end{document}